\pdfoutput=1
\documentclass[letterpaper, 10pt, conference]{ieeeconf}
\IEEEoverridecommandlockouts
\usepackage{amsmath,amssymb}
\usepackage{graphicx}
\usepackage{booktabs}
\usepackage{xcolor}
\usepackage{tikz}
\usetikzlibrary{arrows.meta, positioning, fit, backgrounds}
\usepackage[capitalize]{cleveref}
\usepackage{url}

\newif\ifanonymous
\anonymousfalse

\newcommand{\prompt}[1]{\textit{``#1''}}
\newcommand{\sss}{S$^3$}

\title{\LARGE \bf
Sample, Simulate, Select:\\
Physics-in-the-Loop Text-to-Motion for Humanoids Without Training}

\ifanonymous
\author{Anonymous Authors\thanks{Submission for double-anonymous review.}}
\else
\author{Raphael Memmesheimer and Sven Behnke%
\thanks{All authors are with Autonomous Intelligent Systems, Computer Science
Institute VI, Lamarr Institute for Machine Learning and Artificial Intelligence,
and Center for Robotics, University of Bonn, Bonn, Germany;
{\tt\small memmesheimer@ais.uni-bonn.de}}}
\fi

\usetikzlibrary{shapes.callouts}
\definecolor{promptmag}{HTML}{CC3CCC}   %
\newcommand{\pipelinefig}[3]{%
  \resizebox{\textwidth}{!}{%
  \begin{tikzpicture}[x=1cm, y=1cm, font=\scriptsize,
    box/.style={draw, rounded corners=2pt, align=center, minimum height=3.2em, inner sep=3pt, fill=gray!8, text width=2.45cm},
    img/.style={inner sep=0pt},
    lab/.style={font=\tiny, text=black!55, anchor=north, inner sep=1pt, fill=white},
    arr/.style={-{Latex[length=2mm]}, thick, black!60},
    ev/.style={draw, rounded corners=2pt, align=center, inner sep=3pt, fill=blue!6, text width=6.2cm, font=\scriptsize},
    prompt/.style={rectangle callout, draw=promptmag!60, fill=promptmag!7, line width=0.6pt, rounded corners=3pt,
                   text width=2.25cm, align=center, inner sep=5pt, font=\small\itshape, text=black!85},
  ]
    \ifnum#3=1
      \def\cA{0}\def\cB{3.05}\def\cC{6.15}\def\cD{9.35}\def\cE{12.6}\def\cF{15.65}
    \else
      \def\cA{0}\def\cB{3.6}\def\cC{7.6}\def\cD{7.6}\def\cE{11.6}\def\cF{15.65}
    \fi
    \node[img] (i1) at (\cB, 1.45) {\includegraphics[height=1.85cm]{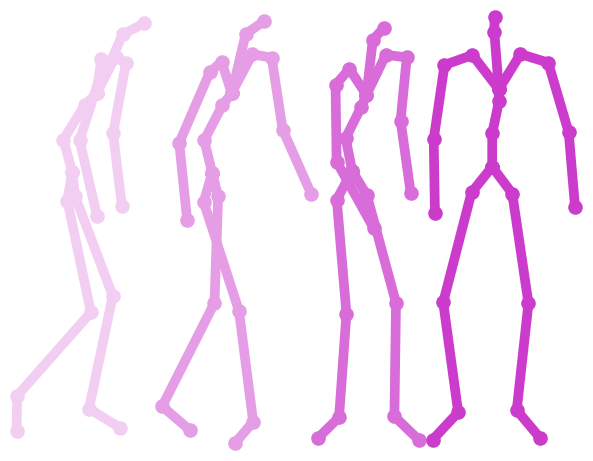}};
    \node[prompt, callout absolute pointer={(\cA-0.95, 0.12)}] (pr) at (\cA, 1.45) {#1};   %
    \node[font=\tiny, text=promptmag!75!black, anchor=south west, inner sep=1pt] at ([yshift=2pt]pr.north west) {text prompt};
    \ifnum#3=1 \node[img] (i2) at (\cC, 1.45) {\includegraphics[height=1.85cm]{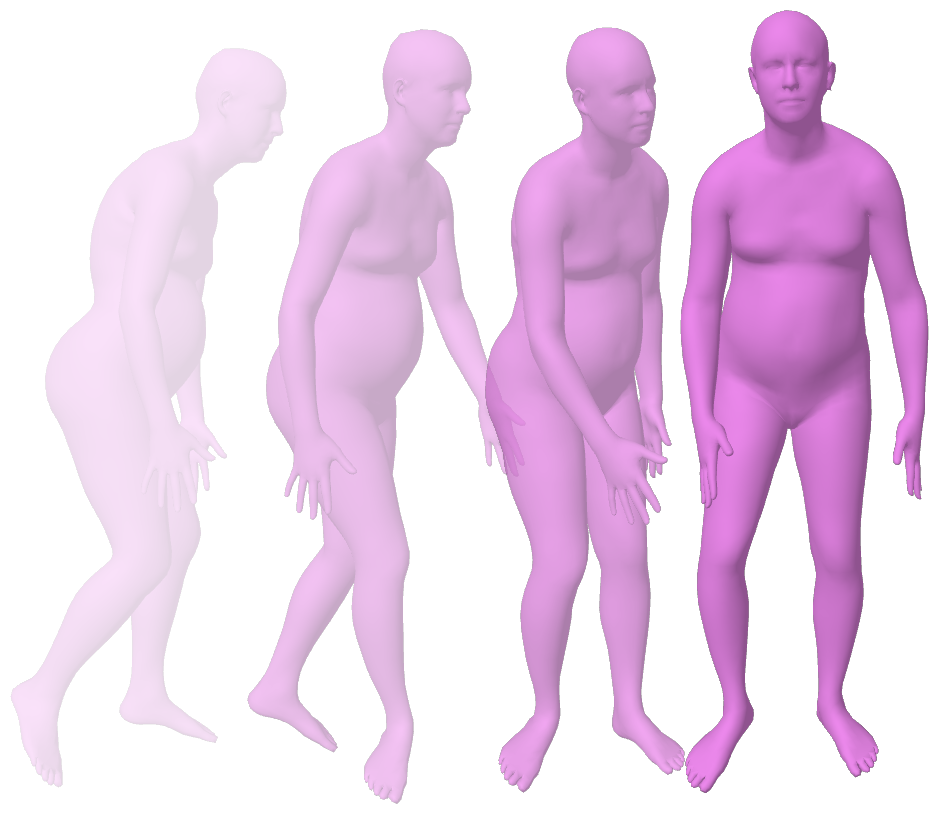}}; \fi
    \node[img] (i3) at (\cD, 1.45) {\includegraphics[height=1.85cm]{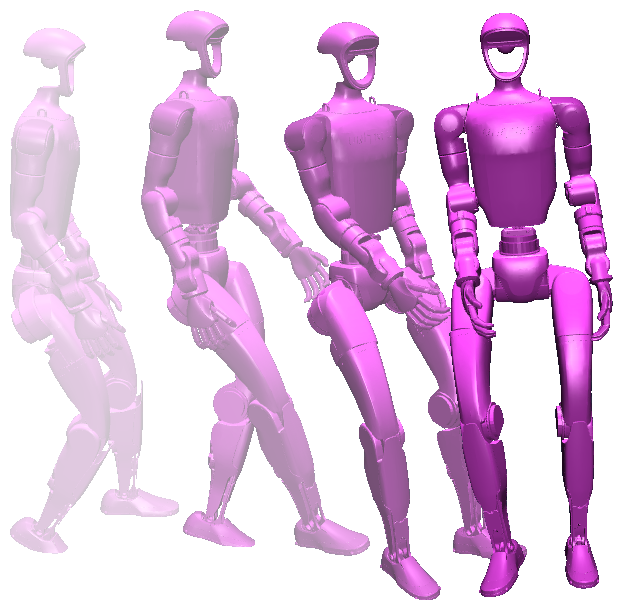}};
    \node[img] (i4) at (\cE, 1.45) {\includegraphics[height=1.85cm]{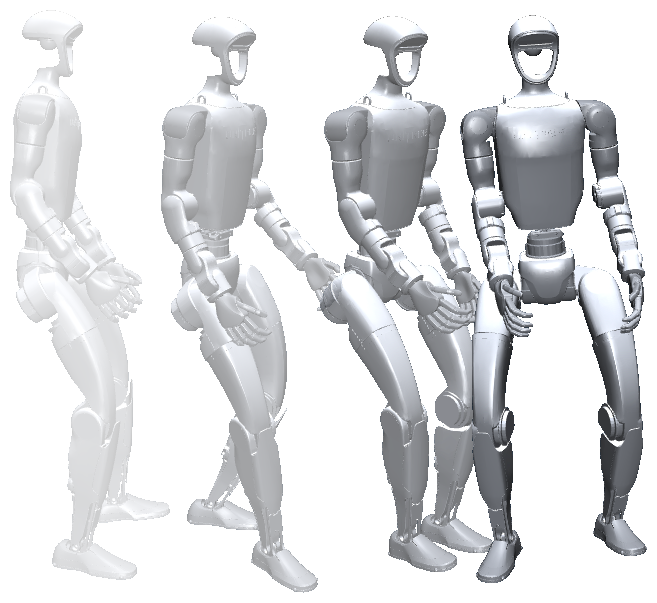}};
    \IfFileExists{images/fig1_p160_f100_real.jpg}{%
      \node[img, draw=black!25, line width=0.4pt] (i5) at (\cF, 1.45) {\includegraphics[height=1.85cm]{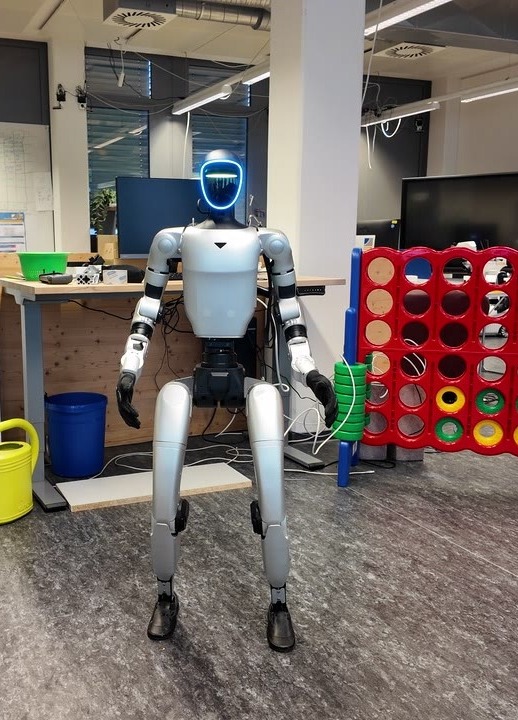}};
      \node[lab] at (i5.south) {real Unitree G1};%
    }{\node[img] at (\cF, 1.45) {\parbox{2.5cm}{\centering\scriptsize\textbf{Real Unitree G1}}};}
    \node[lab] at (i1.south) {joints}; \ifnum#3=1 \node[lab] at (i2.south) {SMPL fit}; \fi
    \node[lab] at (i3.south) {retargeted reference}; \node[lab] at (i4.south) {SONIC, physics};
    \node[box] (b0) at (\cA, -0.75) {\textbf{Prompt}\\HumanML3D test caption};
    \node[box] (b1) at (\cB, -0.75) {\textbf{Sample} $N$\\MoMask~\cite{guo2024momask}, frozen};
    \ifnum#3=1 \node[box, fill=white, draw=black!45, dashed] (b2) at (\cC, -0.75) {SMPL fit\\inspection, GMR ablation}; \fi
    \node[box] (b3) at (\cD, -0.75) {\textbf{Retarget} each\\direction-matching IK};
    \node[box, fill=orange!12] (b4) at (\cE, -0.75) {\textbf{Simulate} each\\SONIC~\cite{luo2025sonic} in MuJoCo};
    \node[box, fill=green!10] (b5) at (\cF, -0.75) {\textbf{Select}\\gate level, then\\lowest tracking error};
    \draw[arr] (b0) -- (b1);
    \ifnum#3=1
      \draw[arr] (b1.east) to[out=0, in=180] (b2.west);
      \draw[arr] (b1.south east) to[out=-70, in=-110, looseness=1.1] (b3.south west);
      \node[font=\tiny, text=black!55] at (\cC, -2.0) {$N$ candidates, all retargeted and simulated};
    \else
      \draw[arr] (b1) -- (b3);
      \node[font=\tiny, text=black!55] at (5.6, -1.85) {$N$ candidates, all retargeted and simulated};
    \fi
    \draw[arr] (b3) -- (b4); \draw[arr] (b4) -- (b5);
    \begin{scope}[on background layer]
      \node[draw=black!25, dashed, rounded corners, inner sep=4pt, fit=(i1)(i4)(b1)(b4)] (frame) {};
      \node[font=\tiny\bfseries, text=black!45, anchor=south east] at (frame.north east) {offline, per prompt};
    \end{scope}
  \end{tikzpicture}}}

\newcommand{\teaserfig}[2]{%
  \begin{tikzpicture}[
    tag/.style={fill=white, fill opacity=0.82, text opacity=1, rounded corners=2pt,
                inner xsep=4pt, inner ysep=3pt, font=\scriptsize, align=left},
  ]
    \node[inner sep=0pt, anchor=south west] (ph) at (0,0)
      {\includegraphics[width=0.9\linewidth]{#1}};
    \node[tag, anchor=south west, font=\small\bfseries\itshape, text width=0.78\linewidth]
      at ([xshift=4pt, yshift=4pt] ph.south west) {``#2''};
  \end{tikzpicture}}

\newlength{\deployh}

\begin{document}
\maketitle
\thispagestyle{empty}
\pagestyle{empty}

\begin{abstract}
Text-to-motion models generate plausible human motion but do not model a
robot's dynamics; whole-body tracking controllers execute robot references
reliably but cannot replan an infeasible one. Recent language-to-humanoid
systems bridge this gap by training. We measure how much of the gap closes
with \emph{no} training at all, by putting the deployment controller itself
in the loop. Sample-simulate-select (\sss) draws $N$ motions per prompt from
a frozen text-to-motion model, retargets each to a Unitree~G1 by
direction-matching inverse kinematics, rolls all of them out under full
rigid-body dynamics with the pretrained SONIC tracking policy, and keeps the
candidate the policy executed best. Because the verifier is the
deterministic simulator itself, \sss\ attains the any-of-$N$ ceiling by
construction; what we measure is where that ceiling lies and what falls
short of it. On 200 stratified HumanML3D test prompts with $N{=}8$, upright
execution rises from 83.5\% to 89.5\% and hardware-gate passes from 33 to
85; on the complete test split (4{,}184 prompts) it rises from 80.5\% to
89.5\%. A kinematic verifier that predicts falls well (AUROC 0.90) recovers
only a quarter of this gain: ranking a prompt's own candidates is harder
than classifying the population. What selection cannot fix is one class,
prompts that lower the pelvis, which a generator trained on retargeted
robot data does execute. We further score the semantic fidelity of the
\emph{executed} motion with the standard text--motion evaluator, with a
real-mocap control that attributes the loss to the robot projection, ablate
the retargeter against GMR (complementary failures: the any-of-8 ceiling
rises to 95.0\% over both), and execute
all 177 gate-selected clips on the real G1: every one completes standing,
with hardware tracking error matching simulation ($r{=}0.94$).
Videos and an interactive browser:
\url{https://raphaelmemmesheimer.github.io/sample-simulate-select/}.
\end{abstract}

\section{Introduction}

Natural language is a convenient interface for specifying humanoid motion:
\prompt{walk forward and wave}, \prompt{sit down}. Two mature lines of work
bring this within reach. Text-to-motion generation for the SMPL
body~\cite{loper2015smpl}, anchored by
HumanML3D~\cite{guo2022humanml3d}, has moved from
diffusion~\cite{tevet2023mdm,zhang2024motiondiffuse,chen2023mld} to token
generators~\cite{zhang2023t2mgpt,jiang2023motiongpt,guo2024momask} trained on
million-clip corpora~\cite{fan2025gotozero,kimodo2026}. In
parallel, reinforcement-learned whole-body trackers reproduce retargeted human
motion on real humanoids~\cite{he2024omnih2o,ji2024exbody2,he2025hover,
chen2025gmt,yin2025unitracker,ze2025twist}, culminating in behaviour foundation
models such as SONIC~\cite{luo2025sonic} that track arbitrary references
zero-shot.

\begin{figure}[t]
  \centering
  \teaserfig{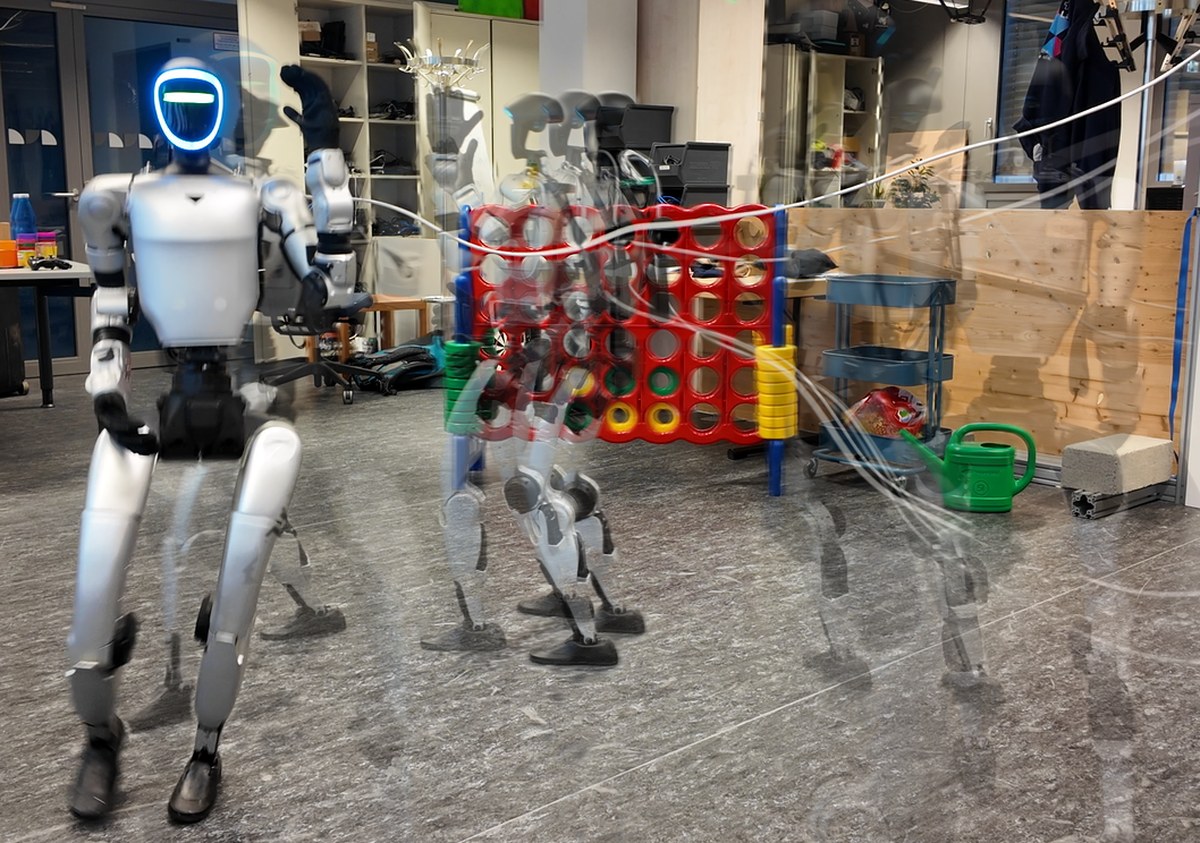}{a man walks in a clockwise circle
  while holding something to his left shoulder}
  \caption{Nothing trained: a frozen text-to-motion model, a pretrained
  tracking policy, a real Unitree G1. One take, instants overlaid.}
  \label{fig:teaser}
\end{figure}

What separates the two is dynamics. A generator trained on human motion has no
notion of the robot's mass distribution, actuator limits or contact
schedule, and a tracker can regularise a reference but not replan it. The
recent wave of language-to-humanoid systems~\cite{mao2025uh1,shao2025langwbc,
li2025roboghost,yue2025rlpf,liu2025humanoidlla,li2026fromw1,xie2026textop,
jia2026echo,cho2026safeflow,cao2026texedo} repairs this inside a model: a
generator trained on retargeted robot data~\cite{mao2025uh1,li2026fromw1}, an
end-to-end language-conditioned policy~\cite{shao2025langwbc}, or a
human-space generator fine-tuned with feedback from a simulated
tracker~\cite{yue2025rlpf,liu2025humanoidlla,cho2026safeflow}.
Each reports success on a real robot; none reports what a frozen composition
of the same public components achieves, so the value added by the training
is never isolated.

\begin{figure*}[t]
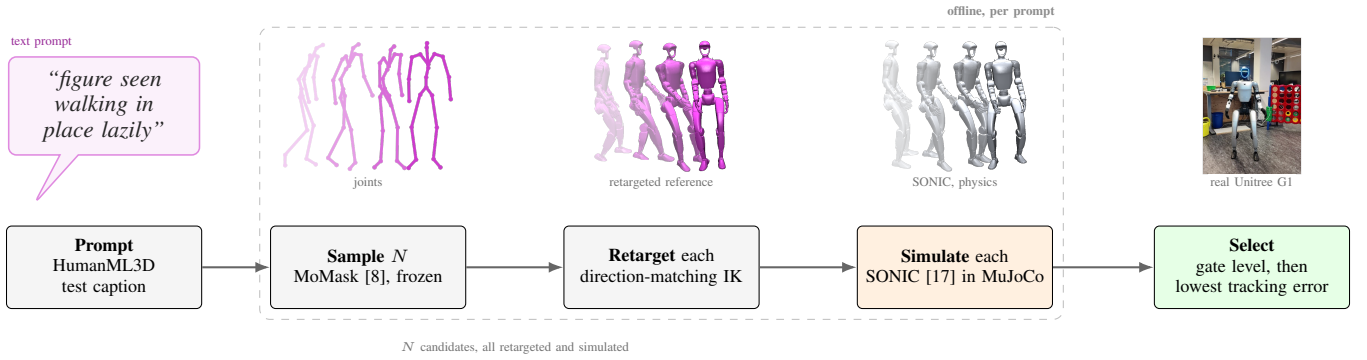

  \centering
  \pipelinefig{``figure seen walking in place lazily''}{fig1_p160_f100}{0}
  \vspace{-20pt}\caption{Sample-simulate-select (\sss). Top: one result (\prompt{figure
  seen walking in place lazily}, a clip later executed on the real G1) in
  the three representations the pipeline passes through, four instants
  overlaid. Bottom: per prompt, $N$ candidates
  are sampled from a frozen text-to-motion model, retargeted to the
  Unitree~G1, rolled out in MuJoCo with the pretrained SONIC policy that also
  runs on the hardware, and the one executed best is selected. Nothing is
  trained.}
  \label{fig:pipeline}
\end{figure*}

This paper measures that value from the other side. Our premise is that the
\emph{deployment controller} is the best available judge of feasibility and
that its cost per candidate is low enough to evaluate all of them: a SONIC
rollout of a five-second clip in MuJoCo takes three seconds on one CPU core,
a MoMask sample about one. Sample-simulate-select (\sss, \cref{fig:pipeline}) therefore
draws $N$ candidates per prompt, retargets and simulates all of them with the
policy that later runs on the hardware, and keeps the candidate the policy
executed best. Nothing is trained, by design: \sss\ is the composition
every trained language-to-humanoid system implicitly claims to improve on
and none has reported, and without it a trained system's success cannot be
attributed to the training rather than to the public components it
inherits. The question is therefore not whether \sss\ beats trained
systems, but how much of the gap they close is \emph{sampling variance
a verifier can exploit} and how much is \emph{motion the generator never
produces feasibly}. With a deterministic simulator as verifier, selection
attains the any-of-$N$ ceiling by construction, so the ceiling itself is
the measurement that separates the two.

Concretely, we contribute (i) \sss, a training-free text-to-humanoid
pipeline whose selection criterion is the physics rollout of the deployment
policy, bounded below by a kinematic-verifier baseline and above by an
any-of-$N$ oracle; (ii) an evaluation protocol on HumanML3D test captions
that stratifies by behaviour category, uses a fall criterion that does not
penalise legitimate crouching, and adds a reference-tracking criterion that
also scores the lying-down and push-up prompts ``upright'' must exclude;
(iii) the measurement of a frozen generator's any-of-8 ceiling, 89.5\% on
the 200 stratified prompts (from 83.5\%) and on the complete test split
(from 80.5\%), of which a kinematic predictor with AUROC 0.90 recovers 1.5
points, and the result that the unrecoverable remainder is a single class,
pelvis-lowering motion, which a generator trained on retargeted robot data
(TEXEDO) does execute under the same verifier; (iv) semantic fidelity of the
\emph{executed} motion on the benchmark's own evaluator with a real-mocap
control, a GMR retargeting ablation, and the execution of all 177
gate-selected clips on the real G1, with hardware tracking error matching
simulation.

\section{Related Work}
\label{sec:related}

\paragraph{Text-to-motion in human space}
HumanML3D~\cite{guo2022humanml3d} fixed the 22-joint, 20\,fps representation
on which token-based~\cite{guo2022tm2t,zhang2023t2mgpt,jiang2023motiongpt} and
diffusion models~\cite{tevet2023mdm,zhang2024motiondiffuse,chen2023mld,
zhang2023remodiffuse,xie2024omnicontrol} are compared; we use MoMask~\cite{guo2024momask}. Scale is the current
frontier~\cite{lin2023motionx,fan2025gotozero,kimodo2026}, but none of these
models represents contact, balance or actuation. Graphics closes that gap for
\emph{simulated} characters with a tracking controller~\cite{peng2018deepmimic,
luo2023phc,tessler2024maskedmimic,tevet2025closd,zhang2026script},
but on a SMPL body whose morphology matches the data; a robot adds retargeting
and hardware.

\paragraph{Retargeting and tracking on humanoids}
Humanoid trackers are built on IK against SMPL keypoints
(H2O~\cite{he2024h2o},
OmniH2O~\cite{he2024omnih2o}, ExBody2~\cite{ji2024exbody2}). Araujo
\emph{et al.}~\cite{araujo2025retargeting} show that retargeting quality
materially affects tracking and release GMR, which we adopt for our
ablation. G1 trackers include
HOVER~\cite{he2025hover}, GMT~\cite{chen2025gmt},
UniTracker~\cite{yin2025unitracker}, TWIST~\cite{ze2025twist} and
BeyondMimic~\cite{liao2025beyondmimic}; SONIC~\cite{luo2025sonic}, a behaviour
foundation model trained on $>$100\,M frames, is the policy we use unchanged.

\paragraph{Language-to-humanoid systems}
\cref{tab:systems} groups recent systems by where language enters and what is
trained. Robot-space generators learn from retargeted data
(UH-1~\cite{mao2025uh1}, Humanoid-LLA~\cite{liu2025humanoidlla},
FRoM-W1~\cite{li2026fromw1}); end-to-end
policies condition on language (LangWBC~\cite{shao2025langwbc}) or on
language-grounded latents (RoboGhost~\cite{li2025roboghost}); physically
aligned generators fine-tune a human-space model with simulated-tracker
feedback (RLPF~\cite{yue2025rlpf}) or guide and
gate a flow model with physics (SafeFlow~\cite{cho2026safeflow}). Modular
pipelines are closest to ours: TextOp~\cite{xie2026textop} streams a diffusion
model into a tracker, ECHO~\cite{jia2026echo} splits generator and tracker
across cloud and edge, and TEXEDO~\cite{cao2026texedo}, the nearest neighbour, samples $N{=}32$
candidates from a generator it trains in G1 joint space on retargeted
AMASS+CLAW and ranks them with a verifier \emph{distilled from SONIC
rollouts}, so that no physics runs at test time. We keep the generator frozen
in human space, use the rollout itself as the verifier (the oracle its
verifier imitates), train nothing, report the any-of-$N$ ceiling
explicitly, and evaluate at a scale none of these report:
physics-validated execution on the complete HumanML3D test split (TEXEDO:
9{,}116 captions of its own split) and 177 distinct prompts on the real
robot (TEXEDO: 30; UH-1: 12).

\begin{table}[b]
  \caption{Recent language-to-humanoid systems: generation space, feasibility
  mechanism, and what is trained.}
  \label{tab:systems}
  \centering\scriptsize\setlength{\tabcolsep}{3pt}
  \begin{tabular}{@{}llll@{}}
    \toprule
    System & Generation space & Feasibility mechanism & Trained \\
    \midrule
    UH-1~\cite{mao2025uh1}             & robot tokens        & data-side            & gen.+policy \\
    FRoM-W1~\cite{li2026fromw1}         & human $\to$ track   & data-side            & gen.+tracker \\
    Humanoid-LLA~\cite{liu2025humanoidlla} & unified vocab.   & RL w/ physics        & LLA \\
    LangWBC~\cite{shao2025langwbc}      & none (end-to-end)   & RL                   & policy \\
    RoboGhost~\cite{li2025roboghost}    & latent              & RL                   & policy \\
    RLPF~\cite{yue2025rlpf}             & human               & RL fine-tuning       & generator \\
    SafeFlow~\cite{cho2026safeflow}     & latent flow         & guidance + gate      & generator \\
    TextOp~\cite{xie2026textop}         & human, streaming    & tracker              & gen.+tracker \\
    ECHO~\cite{jia2026echo}             & human (cloud)       & tracker              & gen.+tracker \\
    PhyGile~\cite{bao2026phygile}       & human, prefixed     & physics prefix       & gen.+tracker \\
    TEXEDO~\cite{cao2026texedo}         & robot, best-of-$N$  & learned verifier     & gen.+verifier \\
    \midrule
    \sss\ (ours)                        & human, best-of-$N$  & deployment rollout   & nothing \\
    \bottomrule
  \end{tabular}
\end{table}

\section{Method}
\label{sec:method}

\subsection{Overview and notation}
Given a prompt $p$, the generator $G$ yields candidates $\mathbf{x}_i = G(p,
\epsilon_i)$, $i=1..N$, each a $T_i\times22\times3$ joint trajectory. The
retargeter $R$ maps each to a robot reference $(\mathbf{q}^\text{ref}_i,
\mathbf{b}_i)$ of joint angles and floating-base pose; the deployment policy
$\pi$ rolled out in simulation produces the achieved trajectory
$\mathbf{q}^\text{phys}_i$ and the outcome features from which a gate
$g(\cdot)\in\{\text{PASS},\text{CAUTION},\text{REJECT}\}$ and a tracking error
$e_i$ are computed. \sss\ returns $\arg\min_i (g_i, e_i)$ in lexicographic
order. Every component below is public and used as released, and all
thresholds were fixed before the study.

\subsection{Sample: frozen text-to-motion}
MoMask~\cite{guo2024momask} with its public HumanML3D checkpoints turns a
prompt into 22 joint positions at 20\,fps; we draw $N$ samples per prompt with
the model's length estimator and default sampling settings, and keep its
foot-contact post-process. Joints are rotated into our $Z$-up convention and
grounded; no scaling is applied.

\subsection{Retarget: direction-matching IK}
\label{sec:retarget}
The source is a position-only 22-joint trajectory, so joint twist and hand
orientation are unobservable; we retarget limb
\emph{directions}. A canonical body frame built from hips and shoulders makes
the subject face $+X$; the root's horizontal translation and full orientation
(pitch and roll damped) drive the robot's floating base. For each
limb chain (thigh, shank, upper arm, forearm on both sides), the goal for
the chain's end body is its anchor body's current position plus the robot's
\emph{own} segment length along the source's unit direction. The result is a
robot-scaled target skeleton that reproduces the human's limb directions
without importing the human's limb lengths. Joint angles follow by damped
least squares over the actuated DoFs, regularised toward the G1's default
stand and the previous frame, clipped to the joint limits and resampled to
50\,Hz.
The map is deterministic and untrained and costs about 15\,s of CPU per
clip. On 1{,}979 AMASS test sequences it reproduces limb orientation to
$4.2^\circ$ on the G1 (elbows worst, $\approx8.5^\circ$, for lack of
forearm articulation);
root-aligned keypoints land 15.5\,cm from the human's, the morphological
floor of joint-space retargeting.

\subsection{Simulate: the deployment policy as verifier}
\label{sec:simulate}
SONIC~\cite{luo2025sonic} is a G1 motion-tracking policy trained on $>$100\,M
mocap frames; we use NVIDIA's released ONNX checkpoint unchanged. References
are name-mapped into its 29-DoF motion-library format and rolled out in
MuJoCo~\cite{todorov2012mujoco} with SONIC's own G1 model, PD actuation and
observation pipeline (50\,Hz control, 200\,Hz physics), reproducing its C++
deployment stack. A clip costs about three CPU-seconds.

\paragraph{Fall criterion}
A height threshold would count every squat as a fall. A rollout is
\emph{fallen} if the pelvis drops below 0.25\,m (on the ground), or its
up-axis tilts more than $60^\circ$ from vertical (toppled), or it sinks more
than 0.30\,m \emph{below the reference pelvis} for longer than 0.5\,s
(collapsed rather than crouched). Crouching references reach 0.16\,m pelvis
height on the G1 and pass this criterion when tracked. Upright is also the
end state we aim for, not only the safe one: a clip that finishes standing
hands over cleanly to the next prompt through the stance blend of
\cref{sec:hardware}, so prompts can be chained. ``Upright'' is
nevertheless a proxy defined on the controller's side, so we also report a
reference-\emph{tracking} criterion that needs no notion of standing: a
rollout is \emph{tracked} if it never sinks 0.30\,m below the reference
pelvis, its pelvis orientation stays within $45^\circ$ of the reference's,
and the mean joint error is below 0.3\,rad. It scores lying, crawling and
push-up prompts that the upright metric must exclude.

\paragraph{Deploy gate}
For hardware we keep a conservative three-level gate on the same rollout:
REJECT if fallen; CAUTION if the first frame deviates more than 0.6\,rad from
the standing pose (the robot drives into it unattended) or any joint rate
exceeds 8\,rad/s; PASS otherwise. Export to the robot additionally requires
a simulated pelvis height of at least 0.5\,m throughout the clip.

\subsection{Select, and what to compare it with}
Per prompt, \sss\ keeps the candidate with the best gate level, ties broken by
the lowest mean joint-tracking error $e_i=\frac{1}{T}\sum_t
\|\mathbf{q}^\text{phys}_t-\mathbf{q}^\text{ref}_t\|_1$, i.e.\ the sample the
policy reproduced most faithfully. Three arms bound it. \emph{Single} is the
first sample (no selection). \emph{Kinematic best-of-$N$} ranks candidates by
a risk score computed from the generator's output alone, the natural
``predict feasibility from kinematics'' baseline: six features (\cref{tab:auroc}) $z$-scored
over the sweep, sign-aligned so that larger means riskier, and averaged. \emph{Oracle} counts a
prompt as upright if any of its $N$ candidates was, the ceiling of any
selection rule. Because the rollout is deterministic and REJECT ranks last,
\sss\ attains this ceiling by construction; the oracle row therefore
measures the generator's headroom at $N$, and the informative comparisons
are the kinematic arm below it and the hardware outcome of what it selects
(\cref{sec:hardware}).

\subsection{Semantic fidelity of the executed motion}
\label{sec:semantic-method}
Selection could trivially favour bland candidates, and retargeting and
tracking could lose the prompt's content. We score every sample with the
HumanML3D text--motion evaluator~\cite{guo2022humanml3d}, the contrastive
encoders behind R-precision and matching score in~\cite{guo2024momask,
tevet2023mdm}, at three stages: the generator's joints, the retargeted
\emph{reference} projected back to 22 SMPL joints by forward kinematics of the
G1, and the \emph{executed} motion projected the same way; the evaluator's
uniform-skeleton step re-imposes human bone lengths. Matching score is the embedding distance to the prompt; R-precision
is whether the prompt is nearest among a batch of 32. To separate what the
evaluator penalises in generated motion from what it penalises in any
robot projection, real HumanML3D test motions are sent through the identical
path as a control (\cref{sec:semantic}).

\subsection{Hardware: batched, gate-screened sessions}
\label{sec:hardware}
Selected clips are exported into SONIC's deployment format and played on the
real G1 through the same C++ binary used for the rollouts.
For safety, every clip is first executed on a gantry that would arrest a
fall; clips that completed cleanly there were afterwards re-executed
without it, in a second pass over a subset. Every exported clip is padded with a 3\,s cosine blend
from the policy's default stance into its first frame and a 2\,s blend back,
so that neither arming nor the end of a clip is a step change on a 35\,kg
robot without fall detection. Clips are exported as sessions of eight,
ordered lowest risk first and re-screened by the gate at export.

\section{Experiments}
\label{sec:experiments}

\subsection{Prompt set and protocol}
We draw 200 captions from the HumanML3D \emph{test} split (4{,}384 motions,
12{,}584 whole-clip captions), taking the shortest caption of each motion
(3--20 words) and stratifying by keyword rules into six behaviour
categories: locomotion (60), turning (30), upper-body (50), squat/bend (30),
ballistic (20), other (10). Two rules make ``upright'' meaningful. Captions
whose motion \emph{requires} leaving the upright state (lying, crawling,
rolling, falling, push-ups, handstands; 128 of the pool) are excluded, since
no upright outcome could satisfy them; captions that legitimately lower the
pelvis (sit, squat, kneel, bend, pick up; 31 of the 200) are labelled
\emph{low-pelvis} and reported separately, and the reference-relative fall
criterion keeps them scorable. The rules are auditable and released;
captions are used verbatim (typos included). All 200 prompts $\times$ 8 samples are generated, retargeted and
rolled out once. The pipeline was then run on the \emph{complete} test split
(4{,}184 usable prompts, 33{,}472 rollouts), reported next to the stratified
set.

\subsection{Main result: the ceiling, and how far selection reaches}

\begin{table}[t]
  \caption{Main result. Four selection arms at $N{=}8$ on the 200 stratified
  prompts, on the same prompts with TEXEDO's trained generator, and on the
  complete test split.}
  \label{tab:arms}
  \centering\scriptsize\setlength{\tabcolsep}{3pt}
  \begin{tabular}{@{}lccrrrr@{}}
    \toprule
    Arm & Upright & 95\,\% CI & PASS & CAUT. & REJ. & $e$ \\
    \midrule
    single (first sample)   & 83.5\% & [77.7, 88.0] & 33 & 134 & 33 & 0.171 \\
    kinematic best-of-8     & 85.0\% & [79.4, 89.3] & 56 & 114 & 30 & 0.163 \\
    \textbf{physics best-of-8 (\sss)} & \textbf{89.5\%} & [84.5, 93.0] & \textbf{85} & 94 & 21 & \textbf{0.138} \\
    oracle (any-of-8)       & 89.5\% & [84.5, 93.0] & 85 & 94 & 21 & 0.135 \\
    \midrule
    \multicolumn{7}{@{}l}{\emph{TEXEDO's trained robot-space generator~\cite{cao2026texedo}, same 200 prompts, same verifier}} \\
    single (first sample)   & 97.0\% & [93.6, 98.6] & 42 & 136 & 22 & -- \\
    physics best-of-8 (\sss) & 98.0\% & [95.0, 99.2] & 101 & 87 & 12 & 0.086 \\
    oracle (any-of-8)       & 98.0\% & [95.0, 99.2] & 101 & 87 & 12 & -- \\
    \midrule
    \multicolumn{7}{@{}l}{\emph{complete test split, 4{,}184 prompts}} \\
    single (first sample)   & 80.5\% & [79.3, 81.7] & 704 & 2646 & 834 & 0.178 \\
    \textbf{physics best-of-8 (\sss)} & \textbf{89.5\%} & [88.6, 90.4] & \textbf{1680} & 2042 & 462 & \textbf{0.134} \\
    oracle (any-of-8)       & 89.5\% & [88.6, 90.4] & 1680 & 2042 & 462 & 0.132 \\
    \bottomrule
  \end{tabular}
\end{table}

\begin{figure}[t]
  \centering
  \includegraphics[width=0.85\linewidth]{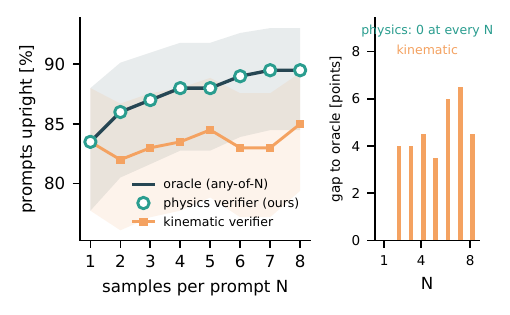}
  \caption{Prompts executed upright vs.\ number of samples $N$. The
  physics verifier (markers) lies on the any-of-$N$ ceiling (line) by
  construction; the kinematic verifier does not improve with $N$. Right: gap
  to the ceiling in points. Shaded: Wilson 95\,\% intervals.}
  \label{fig:bestofN}
\end{figure}

\cref{tab:arms} and \cref{fig:bestofN} contain the central result for the
200 stratified prompts (MoMask, frozen), the same prompts with TEXEDO's
trained generator, and the complete test split; upright is the fall
criterion of \cref{sec:simulate}. Columns: upright rate with Wilson 95\,\% interval,
gate level of the selected sample, mean joint tracking error $e$ [rad]. With a single sample, 83.5\% of prompts
execute upright and 33 pass the hardware gate. Physics selection over eight samples raises this to 89.5\% and 85 PASS
clips. The \sss\ curve lies on the ceiling at every $N$, as it must
(\cref{sec:method}); what \cref{fig:bestofN} measures is the shape of that
ceiling: three quarters of the gain arrive by $N{=}4$ and the curve is flat
from $N{=}7$, so most of the headroom is reached within a small sample
budget. Tracking error of the
selected samples falls from 0.171 to 0.138\,rad, within 0.003 of the best
candidate's. Of the 33 prompts whose first sample fell, 12 are recovered by
selection, and 12 is also the number recoverable with this retargeter: the
remaining 21 fell in all eight samples. The complete test split reproduces this at
20$\times$ the size: 80.5\% single, 89.5\% at the ceiling, 704 to 1{,}680
PASS clips, with the kinematic verifier again halfway (84.5\%) and 438 of
4{,}184 prompts never upright. The conclusions also survive a change of
success criterion: reference tracking (\cref{sec:simulate}) agrees with
``upright'' on 98.4\% of the 33{,}472 samples, and the ceiling under it is
89.6\% (from 79.3\%). Where ``upright'' cannot judge at all, on the 128
test prompts the exclusion rule removes because they demand lying down,
push-ups or kneeling, tracking credits 29.7\% of first samples and 50.0\%
at the ceiling: selection matters most exactly where the generator is
weakest.

\paragraph{Against a trained robot-space generator}
To put the frozen composition next to a trained one, we ran TEXEDO's
released generator~\cite{cao2026texedo}---FSQ-GPT trained in G1 joint space
on AMASS+CLAW retargeted with HumanML3D captions---on the same 200 prompts
with the same $N{=}8$ and the same verifier (\cref{tab:arms}). Training the
generator on retargeted data yields what selection cannot: 97.0\% of first
samples execute (3.6\% of samples fall vs.\ 17.6\%), squat/bend reaches
93\% because its references keep the pelvis higher (executed minimum 0.55
vs.\ 0.35\,m), and selection adds one point (98.0\%, its ceiling) but
lifts PASS clips from 42 to 101. This is not memorisation: 106 of our 200
source motions lie in TEXEDO's training split, yet its rate on the 94 unseen
prompts is the same (97.9\%). The gain in feasibility is not offset by a
loss of semantic fidelity: scored by the same evaluator, its executed
motion is indistinguishable from MoMask's (R@1 0.09 vs.\ 0.08, matching 6.49
vs.\ 6.45) and its kinematic reference slightly lower (R@1 0.11 vs.\ 0.13).

\begin{figure*}[t]
  \centering
  \input{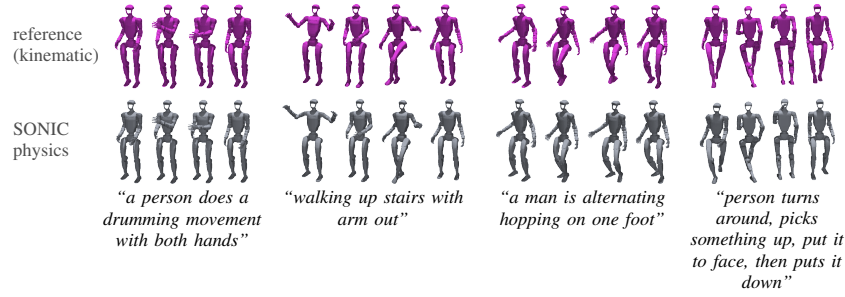}
  \vspace{-5pt}
  \caption{Selected clips that execute well, one per category: retargeted
  reference (top), SONIC rollout (bottom). All four ran on the real G1 and
  completed standing (0.08--0.10\,rad). Squat/bend is the class of
  \cref{fig:failure}, where the success rate is lower but non-zero.}
  \label{fig:good}
\end{figure*}

\subsection{Why a kinematic verifier is not enough}

\begin{table}[t]
  \caption{Predicting a fall from the generator's output alone: per-feature
  and combined AUROC.}
  \label{tab:auroc}
  \centering\scriptsize\setlength{\tabcolsep}{3pt}
  \begin{tabular}{@{}lc@{\hspace{6pt}}lc@{}}
    \toprule
    feature & AUROC & feature & AUROC \\
    \midrule
    trunk lean          & \textbf{0.923} & flight fraction    & 0.597 \\
    pelvis excursion    & 0.791 & peak joint rate    & 0.594 \\
    min.\ pelvis height & 0.752 & start deviation    & 0.580 \\
    \midrule
    combined risk score & 0.899 & & \\
    \bottomrule
  \end{tabular}
\end{table}

Falls are predictable from kinematics: over 1{,}600 samples (282 fell), trunk lean alone reaches AUROC 0.92 and the combined score 0.90
(\cref{tab:auroc}). Yet ranking candidates by
that score gains only 1.5 points (85.0\%), does not improve with $N$
(\cref{fig:bestofN}), and yields 56 rather than 85 PASS clips. The reason is
the difference between classifying and ranking. A feature that separates the
population of fallen from upright samples need not order the eight
candidates \emph{of one prompt}, which share the prompt's semantics and hence
similar lean and excursion. Among the candidates of a prompt whose first
sample fell, the kinematic score prefers a falling one often enough to erase
its advantage, whereas the rollout observes the outcome. This is the
empirical case for physics in the loop rather than a learned
proxy~\cite{cao2026texedo}: a verifier must rank within a prompt, and
population-level accuracy is not evidence that it does.

\subsection{Where selection stops: the failure class}
\label{sec:failure}

\begin{figure}[t]
  \centering
  \includegraphics[width=0.85\linewidth]{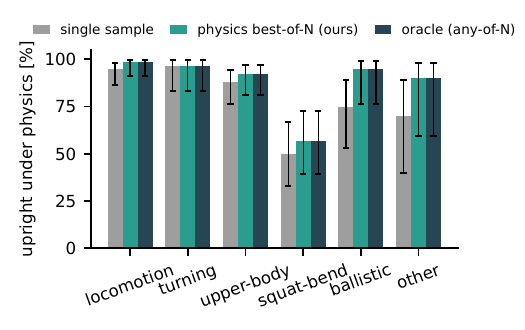}
  \caption{Upright rate per behaviour category. Selection closes the gap in
  every category but squat/bend, where the ceiling is 57\,\%.}
  \label{fig:category}
\end{figure}

\begin{figure*}[t]
  \centering
  \input{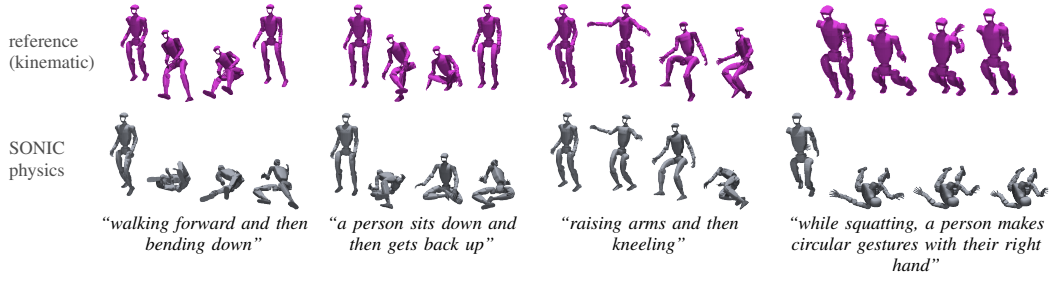}
  \vspace{-5pt}
  \caption{The unrecoverable class: squat/bend prompts on which all eight
  samples fell. Top: the retargeted reference; bottom: the policy follows it
  to the floor.}
  \label{fig:failure}
\end{figure*}

\cref{fig:category} breaks the result down by category. Locomotion (98\%),
turning (97\%) and upper-body prompts (92\%) are essentially solved after
selection. Ballistic prompts, the worst single-sample category apart from
squat/bend (75\%), gain the most and reach 95\%: a jump is frequently
feasible in \emph{some} sample. Squat/bend improves only from 50\% to 57\%,
and that is its ceiling. Restricted to the 169 upright-applicable prompts,
\sss\ reaches 96\% [92, 98]; on the 31 low-pelvis prompts, 55\% [38, 71].
The complete split agrees: 94\% on 3{,}742 applicable, 53\% on 442
low-pelvis prompts. The 21 unrecoverable prompts make the class explicit:
13 are squat/bend, and most of the rest are the same motion under another
label (\prompt{bends down and jumps forward}). A second retargeter (\cref{sec:gmr}) recovers 11
of the 21 in some sample, but the 10 that survive both are the same class:
seven squat/bend, a rise from seated, and a stair climb with no stairs in
the world (\cref{fig:failure}). This is neither sampling variance nor a limit of one
retargeter but motion the frozen generator never produces executably,
exactly what physically aligned
generators~\cite{yue2025rlpf,cho2026safeflow} are trained to
reshape and what TEXEDO's retargeted-data training does reshape: the same
class reaches 93\% there (\cref{tab:arms}).

\subsection{Semantic fidelity: where meaning is lost}
\label{sec:semantic}

\begin{figure*}[t]
  \centering
  \includegraphics[width=0.68\textwidth]{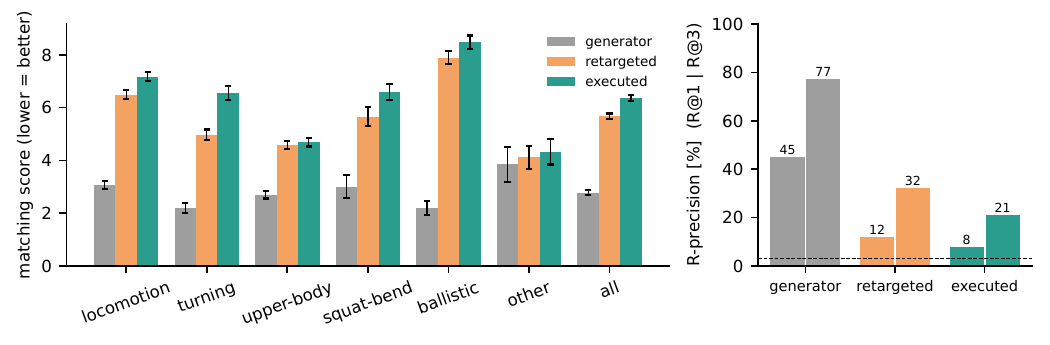}
  \caption{Semantic fidelity at three stages of the pipeline, HumanML3D
  text-motion evaluator on upright samples ($n=1{,}255$). Left: matching
  score (lower is better) per category. Right: R-precision (dashed: chance).
  Most of the loss occurs at retargeting, before any physics.}
  \label{fig:semantic}
\end{figure*}

\begin{table}[t]
  \caption{Semantic fidelity at the generator, after retargeting, and after
  execution.}
  \label{tab:semantic}
  \centering\scriptsize\setlength{\tabcolsep}{2pt}
  \begin{tabular}{@{}lrccrccrcc@{}}
    \toprule
    & \multicolumn{3}{c}{generator} & \multicolumn{3}{c}{retargeted} & \multicolumn{3}{c}{executed} \\
    \cmidrule(lr){2-4}\cmidrule(lr){5-7}\cmidrule(lr){8-10}
    samples & match & R@1 & R@3 & match & R@1 & R@3 & match & R@1 & R@3 \\
    \midrule
    all (1{,}536)          & 2.86 & 0.46 & 0.78 & 5.71 & 0.13 & 0.34 & 6.45 & 0.08 & 0.23 \\
    single-arm picks (192) & 2.82 & 0.47 & 0.79 & 5.71 & 0.15 & 0.37 & 6.46 & 0.09 & 0.21 \\
    \sss\ picks (192)      & 2.84 & 0.43 & 0.78 & 5.70 & 0.11 & 0.33 & 6.43 & 0.07 & 0.23 \\
    \midrule
    real mocap control (68) & 3.27 & 0.26 & 0.62 & 6.86 & 0.07 & 0.19 & 6.80 & 0.04 & 0.18 \\
    \bottomrule
  \end{tabular}
\end{table}

Two questions follow: whether selection sacrifices semantic fidelity for
stability, and how much of the prompt survives the projection onto the
robot. \cref{tab:semantic} answers the first. It reports matching score ($\downarrow$) and
R@1/R@3 ($\uparrow$) for all samples, each arm's picks and the real-mocap
control: the samples \sss\
picks score the same as first samples at every stage (matching 2.84 vs.\
2.82 at the generator, 6.43 vs.\ 6.46 executed); the verifier is blind to
content. For the second, the same motion is scored at the
generator, after retargeting (kinematic reference) and after execution
(\cref{fig:semantic}); eight prompts whose clips are $\le$2\,s, below the
evaluator's minimum length (98 of 4{,}184 on the complete split), are
unscored but kept in the execution results. The generator's samples reach
R@1 0.46 and matching
2.86, consistent with MoMask's published numbers. The retargeted reference,
scored through the robot's forward kinematics before any simulation, already
drops to R@1 0.13 and matching 5.71; executing it costs comparatively little
more (0.08 / 6.45). The complete split gives the same staircase (R@1 0.49
$\to$ 0.13 $\to$ 0.09 over 32{,}688 scored samples). The loss is
category-dependent: upper-body prompts keep the most (R@3 0.40 retargeted,
0.35 executed), whereas ballistic prompts lose everything at retargeting
(R@1 0.00), because the evaluator's features rest on foot contacts and root
velocity that the G1's grounded, shorter-legged reference no longer
reproduces.

A control decides whether this drop indicts the generated motion or the
robot projection: every HumanML3D test motion whose AMASS source we hold
(DFaust, Transitions, SSM; 74 captioned clips, 68 tracked without error) is
rebuilt with HumanML3D's own preprocessing and sent as \emph{real} motion
through the identical path (last row of \cref{tab:semantic}). Real motion
starts lower on the evaluator than MoMask's samples (R@1 0.26 vs.\ 0.46; the
model was trained toward the evaluator's distribution) and falls by the same
mechanism to the same floor (R@1 0.26 $\to$ 0.07 $\to$ 0.04); executed real
and generated motion are indistinguishable to the evaluator. We therefore
read the executed-stage scores as a floor set by the robot's morphology, the
FK proxy skeleton and the evaluator's human training domain, and the
generator-stage scores as the comparison between arms. Caveats: the
evaluator sees our home regularisation and resting wrists, and the control
is small. Only 76.5\% of it
executes upright without selection (DFaust hops, Transitions kicks):
feasibility is a property of the motion, not of its origin.

\subsection{Retargeting ablation: direction IK vs.\ GMR}
\label{sec:gmr}

\begin{table}[t]
  \caption{Retargeting ablation per category: direction-matching IK vs.\ GMR,
  both tracked by SONIC.}
  \label{tab:gmr}
  \centering\scriptsize\setlength{\tabcolsep}{2.5pt}
  \begin{tabular}{@{}lrcccccccc@{}}
    \toprule
    & & \multicolumn{2}{c}{upright, single} & \multicolumn{2}{c}{$e$} & \multicolumn{3}{c}{any-of-8} \\
    \cmidrule(lr){3-4}\cmidrule(lr){5-6}\cmidrule(lr){7-9}
    category & $n$ & dir-IK & GMR & dir-IK & GMR & IK & GMR & both \\
    \midrule
    locomotion  & 60 & \textbf{95.0} [86,98] & 83.3 [72,91] & \textbf{0.135} & 0.159 & 98.3 & 90.0 & 98.3 \\
    turning     & 30 & 96.7 [83,99] & 96.7 [83,99] & \textbf{0.137} & 0.154 & 96.7 & 100 & 100 \\
    upper-body  & 50 & 88.0 [76,94] & \textbf{92.0} [81,97] & \textbf{0.168} & 0.219 & 92.0 & 96.0 & 96.0 \\
    squat/bend  & 30 & 50.0 [33,67] & \textbf{66.7} [49,81] & 0.263 & \textbf{0.217} & 56.7 & 76.7 & 76.7 \\
    ballistic   & 20 & 75.0 [53,89] & \textbf{95.0} [76,99] & 0.200 & \textbf{0.158} & 95.0 & 100 & 100 \\
    other       & 10 & 70.0 [40,89] & \textbf{80.0} [49,94] & \textbf{0.165} & 0.201 & 90.0 & 100 & 100 \\
    \midrule
    all         & 200 & 83.5 [78,88] & \textbf{86.0} [81,90] & \textbf{0.171} & 0.184 & 89.5 & 92.5 & \textbf{95.0} \\
    \bottomrule
  \end{tabular}
\end{table}

We replace our direction-matching IK by the optimisation-based
GMR~\cite{araujo2025retargeting} on the first sample of every prompt
($n=200$). GMR consumes SMPL-X pose, so we fit SMPL to the generated joints
(5--20\,mm joint RMSE) and pass its G1 output through the identical SONIC
conversion and rollout. \cref{tab:gmr} reports first-sample upright rate [Wilson
95\,\% CI], tracking error $e$ [rad], and the any-of-8 ceiling per
retargeter and for their union. Neither retargeter dominates. GMR is 2.5
points ahead overall (86.0 vs.\ 83.5\%, intervals overlapping); ours is
better on locomotion (95.0 vs.\ 83.3\%) and tracks more tightly wherever the
robot stays up ($e$ 0.171 vs.\ 0.184); GMR is better on squat/bend (66.7
vs.\ 50.0\%) and ballistic (95.0 vs.\ 75.0\%), where its references keep a
higher pelvis. The failures are complementary, so the retargeter is a
second axis the verifier can exploit at no training cost: selecting over
both on a single sample reaches 91.0\% (182/200), and over eight samples
each (\cref{tab:gmr}, right) GMR's ceiling is 92.5\% [88.0, 95.4] and the
union 95.0\% [91.0, 97.3] (190/200). On squat/bend the ceiling rises from 56.7 to 76.7\%, so
a third of the prompts unrecoverable under our retargeter are limited by
the retargeting rather than by the generator; the seven that remain
unrecoverable under both are the sits, kneels and deep bends of
\cref{sec:failure}. GMR's
references start 1.10\,rad from the G1's standing pose on average, against
0.69 for our home-regularised map: home regularisation is a safety property
as much as a semantic liability (\cref{sec:semantic}).

\subsection{Hardware sessions}
\label{sec:hardware-results}
\begin{figure*}[t]
  \centering
  \input{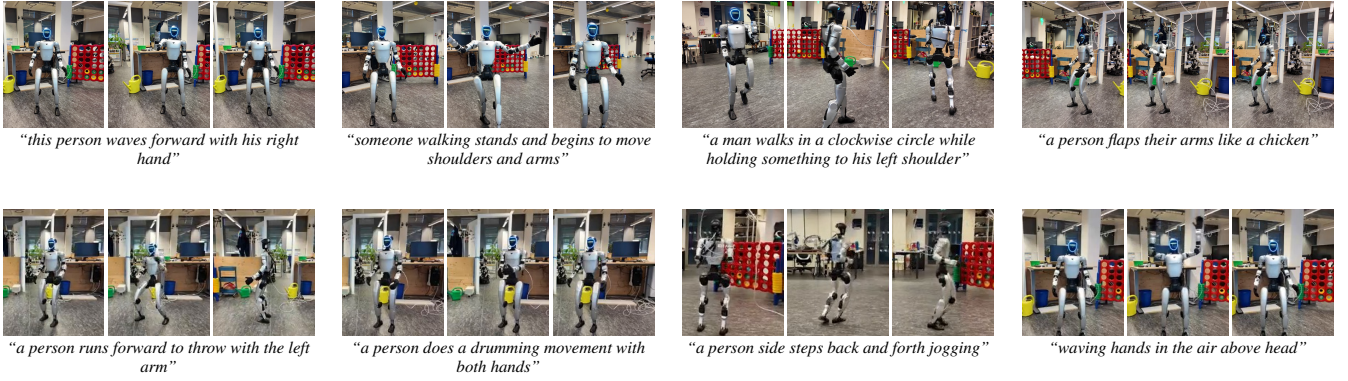}
  \vspace{-17pt}
  \caption{Eight of the 177 gate-selected clips on the real G1, three
  instants each, in time order. All completed standing.}
  \label{fig:hardware}
\end{figure*}
\begin{figure}[t]
  \centering
  \includegraphics[width=0.97\linewidth]{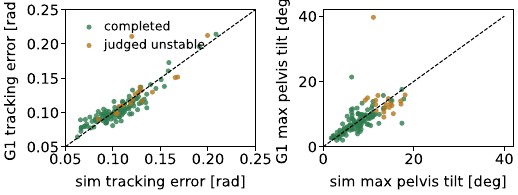}
  \caption{Simulation predicts the hardware: per logged clip, tracking
  error and max pelvis tilt in the rollout vs.\ on the G1 (dashed: identity).}
  \label{fig:hwtracking}
\end{figure}

All 179 clips that pass the gate (85 PASS, 94 CAUTION) were exported
lowest risk first as 23 sessions of up to eight clips; two kneeling clips
with a simulated pelvis height of 0.46\,m fell under the 0.5\,m export
floor, leaving 177. All 23 sessions were executed on the G1 over two days,
one trial per clip (\cref{fig:hardware}). \textbf{All 177 clips completed with the robot
standing}: no fall, no abort, PASS 85/85 and CAUTION 92/92, a Wilson 95\,\%
lower bound of 0.98 on the hardware success of gate-selected clips. The
operator judged 21 \emph{unstable} (visible wobble, or a motion not executed
as prompted: 15 of the 19 ballistic clips never left the ground), all of
which recovered without gantry load. A subset judged safe from this first
pass was then re-executed with the gantry removed, and completed standing
again. The controller logs of 134 clips
(\cref{fig:hwtracking}) show that the simulation is a faithful verifier:
mean joint tracking error on the robot is 0.114\,rad against 0.115\,rad in
simulation for the same clips (Pearson $r=0.94$), and the clips judged
unstable are the ones both sides flag (error 0.150 vs.\ 0.108\,rad, maximum
pelvis tilt $14^\circ$ vs.\ $8^\circ$, in simulation as on the robot). The
rollout also predicts instability per clip: its maximum pelvis tilt
separates the 21 unstable from the 156 stable clips with AUROC 0.94, as
well as the tilt measured on the robot itself (0.92) and better than the
kinematic risk score (0.81) or the simulated tracking error (0.79). The
two clips the export floor had refused were tried with the floor lowered to
0.45\,m: both lowered the body unstably and did not recover, so the floor
sits where the hardware fails.

\section{Discussion}

\paragraph{Scope and limits of rollout-based selection}
Selection with the deployment policy realises the generator's any-of-$N$
ceiling at a few CPU-seconds per candidate and raises the hardware-ready
yield by 2.6$\times$, but cannot create motion the generator does not
produce. The head-to-head with TEXEDO's generator makes the split
quantitative: a generator trained on retargeted robot data lifts first-sample
execution from 83.5 to 97\% and closes squat/bend, while selection lifts the
frozen generator from 83.5 to its 89.5\% ceiling and the trained one by one
point. Training raises the feasibility of the samples themselves, selection
recovers the feasible ones already present, and both leave a remainder that
no verifier can remove. A learned verifier can improve on the rollout only in evaluation cost,
the trade-off TEXEDO's distilled verifier makes.

\paragraph{Why keep generation in human space}
Robot-space generators embed the robot in the model. Ours transfers to other
humanoids through a chain definition, attributes a failure to the prompt,
the generator or the robot, and scores the executed motion on the human
benchmark's own scale (\cref{sec:semantic}).

\paragraph{Threats to validity}
Two generators, one retargeter family, one controller, one robot: the
numbers bound these compositions, not the design space. Simulation is
SONIC's own MuJoCo model without system identification, so a PASS is
evidence about the deployment stack rather than a guarantee; the hardware
sessions are the check. Category and upright labels come from
keyword rules, and ``upright'' is a proxy inherited from tracking
evaluation; the tracking criterion is the general one and agrees with it
where both apply. $N{=}8$ is a budget, not a limit.

\paragraph{Limitations and outlook}
The pipeline is offline and open-loop; TextOp~\cite{xie2026textop} shows
what streaming adds, and the verifier is
fast enough to run inside such a loop. Neither the generator nor the verifier
senses the scene: no visual or other exteroceptive measurement enters the
loop, and no object interaction is modelled. The rollout is the robot on flat
ground with no props, so a prompt presupposing a chair, a step or a handled
object is scored as free-space motion---one reason the pelvis-lowering class
fails.

\section{Conclusion}
Sampling several candidates from a frozen text-to-motion model, simulating
all of them with the whole-body policy that will run on the robot, and
keeping the one it executed best raises upright execution from 83.5\% to
89.5\% on 200 HumanML3D test prompts and from 80.5\% to 89.5\% on the
complete test split, the frozen generator's any-of-8 ceiling, and all 177
gate-selected clips completed standing on the real G1, with no training
anywhere. A kinematic verifier that predicts falls well still cannot rank
candidates well. What remains infeasible lowers the pelvis, and a generator
trained on retargeted robot data closes exactly that class. That split, the
semantic-fidelity floor of the robot projection, and the complementarity of
two retargeters are the reference points against which trained
language-to-humanoid systems should be measured.

\bibliographystyle{IEEEtran}
\bibliography{references}

\end{document}